%% file: 0_main.tex
\documentclass[letterpaper]{article}
\usepackage{ijcai19}
\usepackage{times}  
\usepackage{helvet}  
\usepackage{courier}  
\usepackage{url}  
\usepackage{graphicx}  
\usepackage{color}
\usepackage[table]{xcolor}
\usepackage{subfig}
\usepackage{amsmath}
\usepackage{amsfonts}
\usepackage{comment}
\usepackage{booktabs}
\usepackage{wrapfig}
\newcommand{\dvec}{f^{d2v}}
\newcommand{\dembi}{\mathbf{z}_i}
\newcommand{\loci}{\mathbf{c}_i}
\newcommand{\nightonly}{\textbf{Nightlight-Only}}
\newcommand{\wikipedia}{\textbf{Wikipedia Embedding}}
\newcommand{\multimodal}{\textbf{Multi-Modal}}

\title{Predicting Economic Development using Geolocated Wikipedia Articles}

\author{
*Evan Sheehan$^1$\footnote{Co-first Author},
*Chenlin Meng$^1$,
*Matthew Tan$^1$,
Burak Uzkent$^1$,
Neal Jean$^1$,
David Lobell$^2$,
Marshall Burke$^2$,
Stefano Ermon$^1$
\\
\affiliations
$^1$Department of Computer Science, Stanford University \\
$^2$Department of Earth Systems Science, Stanford University \\
}

\begin{document}
\maketitle
%

%
\begin{abstract}
\input{0_abstract}
\end{abstract}

%


\input{0_intro}

\input{0_datasets}

\input{0_methods}

\input{0_results_poverty}

\input{0_interpretability}

\input{0_results_education}

\input{0_related}
\input{0_conclusion}



\bibliographystyle{named}
\bibliography{acmart}

\end{document}

%% file: 0_abstract.tex
Progress on the UN Sustainable Development Goals (SDGs) is hampered by a persistent lack of data regarding key social, environmental, and economic indicators, particularly in developing countries. For example, data on poverty  --- the first of seventeen SDGs --- is both spatially sparse and infrequently collected in Sub-Saharan Africa due to the high cost of surveys. 
Here we propose a novel method for estimating socioeconomic indicators using open-source, geolocated textual information from Wikipedia articles. We demonstrate that modern NLP techniques can be used to predict community-level asset wealth and education outcomes using nearby geolocated Wikipedia articles. When paired with nightlights satellite imagery, our method outperforms all previously published benchmarks for this prediction task, indicating the potential of Wikipedia to inform both research in the social sciences and future policy decisions.

%% file: 0_intro.tex
\section{Introduction}

\input{00_wiki_dataset.tex}

The number one goal of the United Nations Sustainable Development Goals is to "end poverty, in all its forms, everywhere." While much has been done to achieve this objective, there still remain vast regions of the world where extreme poverty continues to be a persistent and endemic problem. Progress is hampered by a stubborn lack of data regarding key social, environmental, and economic indicators that would inform research and policy. Many nations lack the governmental, social, physical, and financial capabilities to conduct large-scale data gathering operations and costly on-the-ground surveys to identify and distribute aid to the most needy communities~\cite{sahn2000poverty}.
Unfortunately, this data scarcity is typical of other SDGs as well: healthcare and mortality data \cite{kahn2007research}, infrastructure and transportation statistics, economic well-being and educational achievement information, food security \cite{antwi2012mapping} and wealth inequality assessment, among many others. This lack of detailed and consistent information contributes to delayed or suboptimal financial and physical responses from both regional governments and international aid organizations. 

There have been numerous attempts to remedy this shortage of socioeconomic information through the combination of machine learning with cheap, globally available data streams such as social media or remotely sensed data. 
In particular, with regard to healthcare data, \cite{signorini2011use} use Twitter as an information-rich source from which to track and predict disease levels and the public's concerns regarding pathogens, while \cite{carneiro2009google} pursue a similar task via the analysis of Google trends. \cite{pulse2014mining} mine the tweets of Indonesian citizens in order to understand food shortages as well as analyze food security and predict food prices on a granular level. 
\cite{nikfarjam2015pharmacovigilance} also utilize social media posts to track adverse reactions to drugs with the hope of increasing the scale of data available in the healthcare space. Finally, \cite{blumenstock2015predicting} use mobile phone metadata to attempt to predict poverty levels for data-sparse regions of Rwanda, and \cite{xu2016cross} successfully perform detailed traffic prediction using the open-source geospatial dataset OpenStreetMap (OSM) \cite{haklay2008openstreetmap}.

In this paper, we propose the use of Wikipedia as a previously untapped source of data to estimate socioeconomic indicators at very high spatial resolution. Our approach is motivated by both the sheer scale and information-richness of Wikipedia (see Fig. \ref{fig:Wiki Dataset} for a spatial distribution of articles), as well as the ease with which its geospatial nature can be exploited. Indeed, in the English Wikipedia alone, there are over 5.7 million entries, each of which averages 640 words in length. Of these, 1.02 million are geolocated (i.e., associated with a physical coordinate) (Fig. \ref{fig:Wiki Dataset}), allowing them to be used for spatial socioeconomic indicator prediction. Surprisingly, many developing regions of the world contain high concentrations of geolocated articles, with 50,000 for the continent of Africa alone. By providing detailed textual information about locations and entities in an area (e.g., articles about dams, schools, hospitals), we view the articles as data-rich proxies representing the regions around them~ \cite{sheehan2018learning,uzkent2019learning}. In principle, articles like these contain socioeconomic information, though it is often more qualitative than quantitative. Because much of this qualitative data is in unstructured article bodies, and not in infobox/structured formats, it is difficult to extract --- this difficulty has prevented Wikipedia from being integrated fully into social science research.


As such, we propose leveraging recent advances in NLP to extract information from free-text Wikipedia articles. We first map geolocated articles to a  vector representation using the $\textit{Doc2Vec}$ method \cite{le2014distributed}. We then use the spatial distribution of these embeddings to predict socioeconomic indicators of poverty, as measured by ground-truth survey data collected by the World Bank.
Using the latent embeddings of nearby geolocated Wikipedia articles to predict the wealth level of geographic locales, we demonstrate that, on its own, Wikipedia is a strong socioeconomic predictor of poverty across 5 Sub-Saharan African nations: Malawi, Nigeria, Tanzania, Uganda, and Ghana. 
We further extend our models to incorporate local information about nightime-light intensity as measured by satellites.
We then show that this multi-modal approach combining textual information and remotely sensed data significantly improved state-of-the-art results for the task of poverty/economic well-being prediction across these same 5 countries. We achieve Pearson's $r^2$'s of .64, .70, .71, .76, and .76, respectively, as compared to current nightlight state-of-the-art results of .55, .69, .58, and .66 (Ghana not included in previous study). 
Additionally, our approach is able to provide reliable predictions of education-related outcomes, indicating that it can generalize to different types of socioeconomic indicators. 
We believe the varying data sources and data types in this multi-modal approach provide our models with much more descriptive features. To the best of our knowledge, there has not heretofore been any work done utilizing multi-modal data for poverty prediction.

%% file: 00_wiki_dataset.tex
\begin{figure*}
\subfloat{
        \includegraphics[width=0.37\textwidth]{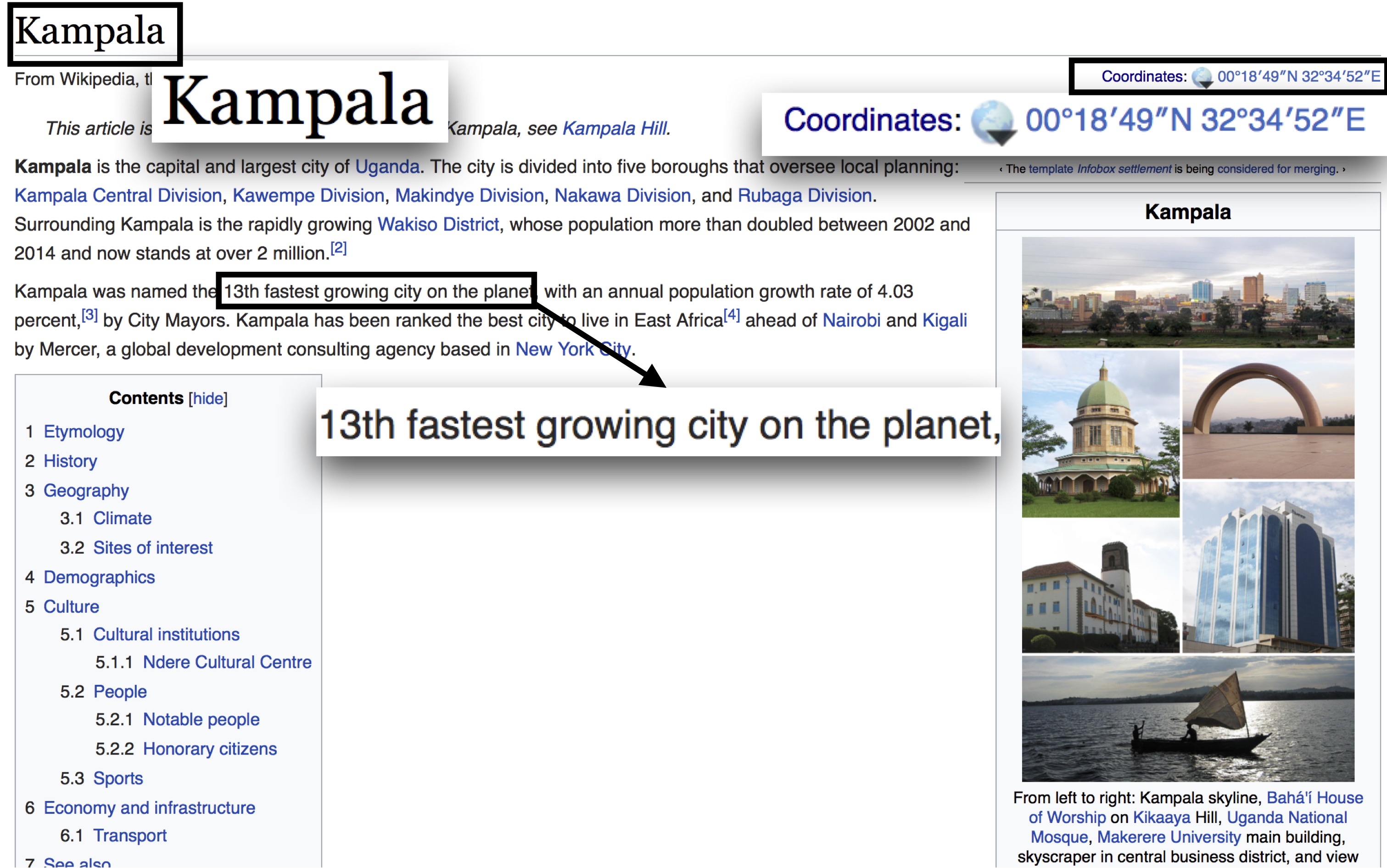}
        \label{fig:Sample Wiki}}
\hspace{1in}
\subfloat{
        \centering
        \includegraphics[width=0.37\textwidth]{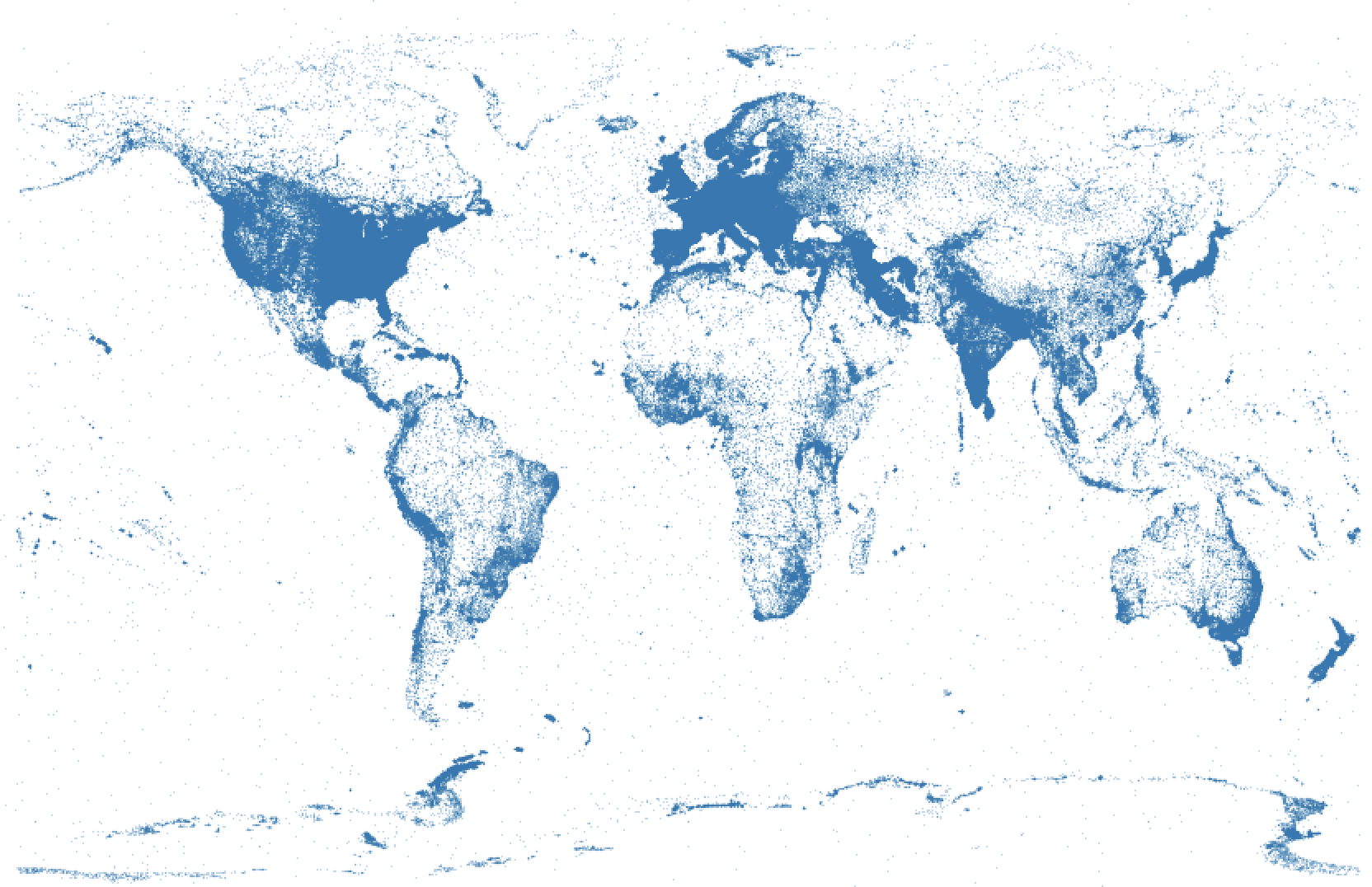}
        \label{fig:Global Wiki}}
    \caption{\textbf{Left:} Example of a geolocated Wikipedia article. Articles such as this contain a wealth of information relevant to economic development. \textbf{Right:} Global distribution of geolocated Wikipedia Articles. Note that there is no overlayed map, and the shape of the continents arises naturally from the spatial distribution of articles.}
    \label{fig:Wiki Dataset}
\end{figure*}

%% file: 0_datasets.tex
\section{Datasets}

In this section, we describe the ground truth data for the indexes used to measure poverty and education as well as the collection of Wikipedia articles and nightlight images.

\input{00_ground_truth_dataset.tex}


\subsection{Socioeconomic data}
Our ground truth dataset consists of data on asset ownership from the Demographic and Health Surveys (DHS)~\cite{DHS} and Afrobarometer~\cite{AFRO}. DHS is a set of nationally-representative household surveys conducted periodically in most developing countries, spanning 31 African countries. Afrobarometer (Round 6) is a survey across 36 countries in Africa, assessing various aspects of welfare, education, and infrastructure quality.
While data from the DHS and Afrobarometer has been used as training data in multiple machine learning applications~\cite{Jean790,blumenstock2015predicting,Xie2016TransferLF,Oshri:2018:IQA:3219819.3219924}, it is both costly to create and sparse in its coverage. 
Typically, the door-to-door surveys used in DHS are conducted approximately every five years, resulting in poor temporal resolution. Each survey also yields only a few thousand data points per country, even for nations with tens of millions of residents. This means the data captures only a tiny fraction of the population, often leading to entire regions within countries being left out of the survey. Additionally, these surveys are not conducted in every country, and destabilized and dangerous regions of the globe can be excluded due to the hazard presented to interviewers.

From the available DHS survey data, we compute an Asset Wealth Index (AWI) as the first principal component of survey responses to questions about asset ownership (e.g. ownership of bicycles, cars, etc.); this approach is a common one for building a measure of wealth from survey data ~\cite{filmer1999effect,smits2015international}. To make this index comparable across all our African countries, we compute the PCA on the data pooled across all countries and survey years, yielding a normalized measure scaled between approximately -2 and 2 (see Fig.~\ref{Ground Truth Dataset}). We aggregate household-level AWI estimates to the village level -- the level at which our latitude and longitude data are available -- yielding a training set where labels correspond to a wealth index $y_{i}$ for each lat/lon coordinate $c_{i}$ where the survey was taken. To protect privacy, the recorded coordinate $c_{i}$ for each datapoint provided by DHS contains up to 5 km of added noise in rural areas and 2 km of noise in urban areas. This limits the usable spatial resolution of the data. A visualization of this effect is in Fig.~\ref{Ground Truth Dataset}, right.

To test the effectiveness of our approach on different outcomes, we also use an education index directly from Afrobarometer, capturing the local level of education in a community. The use of Afrobarometer, a survey conducted by a different organization and with a different methodology, allows us to further test the generality of our approach.


\subsection{Geolocated Wikipedia Corpus}
Our input dataset consists of a corpus of geolocated Wikipedia articles $D$. To obtain it, we parsed the June 2018 Wikipedia article dump into its constituent documents and then extracted all those that were geolocated (i.e., associated with a specific latitude and longitude). This process netted over 1 million geolocated articles globally (see Fig.~\ref{fig:Wiki Dataset} for this distribution as well as an example article). Of note is that throughout Africa there are roughly 50,000 such articles. 
These geolocated articles are about topics encompassing many categories, including infrastructure (i.e., bridges and roads), populated places, companies, and others. While some articles contain structured information in their infoboxes, many simply contain unstructured text data in their article bodies. 

\subsection{Nightlights Imagery}

In addition to learning textual features from geolocated Wikipedia articles, we utilize (5 km $\times$ 5 km) nightlights images compiled by \cite{nips_2017_workshop_stefano}. 
These images are composed of high-resolution night-time imagery from VIIRS \cite{VIIRS} and are of shape $(255,255)$. For each point in the AWI data that we train on, we obtain a nightlights image centered on its coordinate $c_i$. The size of these images was set to (5 km $\times$ 5 km) due to the (maximum) 5 km noise that each location $c_i$ has.
Nightlights information has been shown to be highly correlated with economic development~\cite{Xie2016TransferLF}. However, nightlights are less effective in the poorest quantiles of the population~\cite{Jean790}. In fact, there are numerous locations that are completely dark at night, yet exhibit significant variation in terms of wealth index (see Fig.~\ref{Prediction}). This motivates the use of alternative covariates for prediction, such as Wikipedia articles.




%% file: 00_ground_truth_dataset.tex
\begin{figure*}[t]
    \centering
    \subfloat{
        \includegraphics[width=0.3\textwidth]{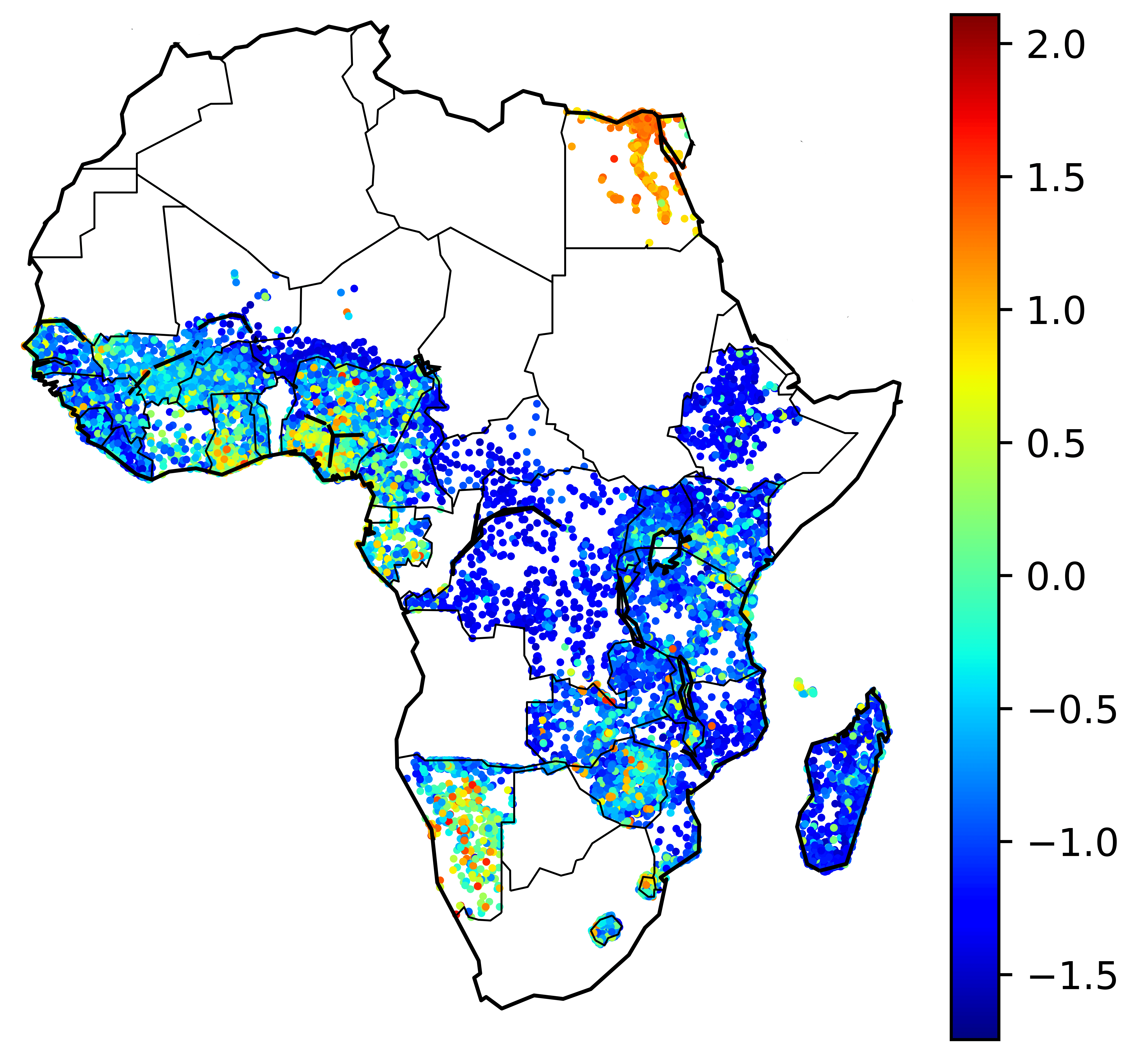}
        \label{Ground_truth_africa}}
    \hspace{1.5in}
    \subfloat{\includegraphics[width=0.3\textwidth]{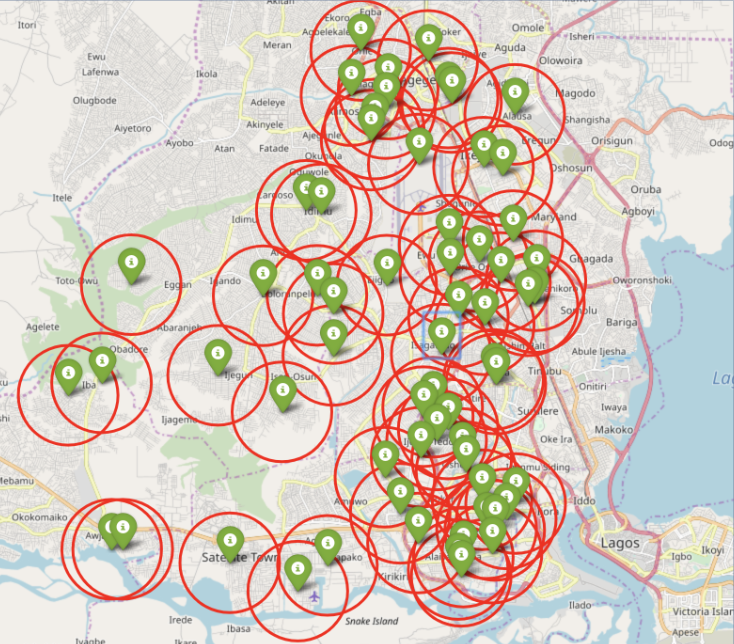}
        \label{Dataset Jitter}}
    \caption{\textbf{Left:} Visualization of ground-truth Asset Wealth Index (AWI) data. Higher values (red) indicate wealthier communities. \textbf{Right:} Jitter in Lagos, Nigeria. Coordinates have up to a 2 km jitter radius in urban areas and 5 km in rural ones.}
    \label{Ground Truth Dataset}
\end{figure*}

%% file: 0_methods.tex
\section{Methods}

We consider the task of predicting the wealth level $y_{i}$ at various locations $c_{i}$ (represented by a latitude, longitude pair) using geolocated Wikipedia articles from a suitable neighborhood of $c_{i}$. In order to accomplish this, we use the AWI data described in the previous section as ground truth to train and evaluate our model.
Processing these Wikipedia articles poses a large challenge due to the unconstrained textual data they contain. Much of the data is not stored in structured quantitative ``infobox'' format, but instead in qualitative and textual format in the article bodies. 
As an example, one can consider the articles on \textit{Kampala}
and \textit{Masaka}
, two different cities in Uganda. 
In these articles, the infobox data contains only population totals, weather data, elevation, and a few other easily obtained metrics. However, in the unstructured body text, much richer qualitative data is available. For instance, the Kampala article describes construction of a new light rail system in the city, as well as efforts to relocate heavy industry to ease traffic. 
The Masaka article discusses the destruction of the city in the 1980's during a civil war and also lists the most common economic activities of its citizens. These types of facts are likely strongly correlated with the regions' economic well-being, but the data is stored as unstructured text. At the same time, many articles (e.g., about historical events) contain information that is likely not relevant for our predictive task --- this makes extracting the relevant information an even more difficult task. 

An additional challenge is posed by the biased and censored nature of our Wikipedia corpus. The spatial distribution of articles, their length, and their content vary widely across countries and even across regions within the same country. The fact that there are no articles on hospitals in a region obviously does not mean there is no hospital in that region. Similarly, even if an article is present, it might or might not mention aspects that would be useful for our task. Unfortunately, even the probability that relevant facts (e.g., hospital presence) are mentioned in Wikipedia appears to vary significantly across regions. This makes statistical inference much more difficult. This challenge is of course shared by many other sources of social media data, e.g., Twitter messages. However, it does not arise when using a data source such as nightlights, which are collected uniformly and consistently\footnote{Bias due to geography is corrected for using physics-based models.} across the globe.

Following the proposed framework 
, we seek to address the following questions: 
(1) Can we leverage the information in geolocated Wikipedia articles to predict economic indicators, despite the unstructured nature of free text and the inherent bias in the corpus?
(2) How does the quality of the predictions compare to conventional approaches based on remote sensing data?
(3) Is the information from Wikipedia complementary, i.e., can we develop a multi-modal approach utilizing Wikipedia articles  alongside nightlights imagery?



\subsection{Learning Representations of Wikipedia Articles}

The problem of learning suitable representations for unstructured text data has received increasing attention recently \cite{pang2016text,wang2019learning,lei2015predicting}. 
Yet many common methods, such as TF-IDF, Bag-of-Words, and \textit{Word2Vec} \cite{mikolov2013distributed}, struggle to capture long-range semantic dependencies between words.
The \textit{Doc2Vec} approach~\cite{le2014distributed} remedies these shortcomings by learning robust representations of not only words, but also paragraphs and documents.  \textit{Doc2Vec} captures long-range word dependencies, word orderings, and holistic semantics of words in a document, and has been used in recent studies \cite{wu2018image,ramisa2018breakingnews} that leverage both textual and visual data. Additionally, it has been used on Wikipedia articles to assess the quality of articles \cite{dang2016quality,dang2017end}, with an in-depth empirical analysis presented in~\cite{lau2016empirical}. 
Similarly to these works, we use \textit{Doc2Vec} to embed a Wikipedia article, $D_{i}$, into a $p$-dimensional distributed representation suitable for our downstream task. 

To train \textit{Doc2Vec}, we preprocess the body of 1.02 million geolocated Wikipedia articles, $D$, by tokenizing and lowercasing text and removing extraneous punctuation.  This corpus of articles is then fed into a \textit{Doc2Vec} model that utilizes the Paragraph Vector - Distributed Bag of Words (PV-DBOW) training method \cite{le2014distributed} at the suggestion of \cite{lau2016empirical} and others. The model then learns to project the articles to $p$-dimensional vector representations. For this task, we choose a window size of 8, a dimensionality $p$ of 300, and train the model for 10 epochs --- these hyperparameters are similar to those suggested by~\cite{lau2016empirical} as being optimal for similar tasks. We denote the function that is parametrized by the $\textit{Doc2Vec}$ model and returns the $p$-dimensional representation for article $D_i$:
\begin{equation}
    \dembi = \dvec(D_{i}).
\end{equation}

\subsection{Wikipedia Embedding Model}
 
We first propose a model to estimate AWI poverty scores $y_i$ utilizing only textual information in Wikipedia articles in a neighborhood of $\mathbf{c}_i$. 
Suppose that, for a specific country, we have $M$ points where we have DHS ground survey data. 
Each of these points corresponds to a location $\loci = (c_i^{lat}, c_i^{long})$ where $i\in \{1,...,M\}$. 
For each $\loci$  we iterate through all the geolocated Wikipedia articles and find $N$ Wikipedia articles 
$D_{i}^1,...,D_{i}^N$
, each with corresponding location 
$(\mathbf{l}_i^1, \cdots, \mathbf{l}_i^N)$
, 
that are closest to $\mathbf{c}_{i}$ w.r.t. Euclidean distance.
We concatenate the $N$ closest article embeddings $(\dvec(D_i^1), \cdots, \dvec(D_i^N))$, each of which is a vector in $\mathbb{R}^{300}$.
Additionally, we compute their Euclidean distances (km) to $\loci$, $(\| \mathbf{l}_i^1-\loci\|,\cdots,\|\mathbf{l}_i^N-\loci\|)$, and use it as an additional feature. This is to capture the potential lack of articles nearby and allow the model to adjust the predictions as necessary.
 Thus, the final proposed text embedding can be formulated as:
\begin{align}
z_{i}^{t} = 
(\dvec(D_i^1), \cdots, \dvec(D_i^N), \|\mathbf{l}_i^1-\loci\|,\cdots,\|\mathbf{l}_i^N-\loci\|) 
\label{eq:dist}
\end{align}
This provides us with a vector in $\mathbb{R}^{300\times N + N}$ \
for each $\loci$.
The combination of textual and distance embeddings are then passed through a neural network consisting of a 3 layer Multi-Layer Perceptron (MLP) with sigmoid and ReLU activations. 
Lastly
we perform regression to predict DHS values (e.g., wealth) for the ground truth point $\loci$.
A schematic representation of the model is shown in Fig.~\ref{fig:wikipedia_only_model}. 

We empirically determine $N$ (the number of closest articles we consider) in our results section (see Fig.~\ref{Article Iteration Graph}). Fig.~\ref{Article Distances} displays the distances to the closest and 10th closest articles for each wealth coordinate $\mathbf{c}_{i}$. Based on these figures, we find that majority of the articles are within 100 km of their corresponding data point.


\input{00_article_distances.tex}

\subsection{Multi-Modal Architecture}
Following our textual only model, we propose a novel multi-modal joint architecture that utilizes nightlight imagery and textual information from Wikipedia. 
For each point $\loci$, we utilize our Wikipedia article 
features from Eq.~\ref{eq:dist} as in the previous section. 
At the same time, we obtain the (5 km $\times$ 5 km) VIIRS nightlight image $x_{i}$ centered at $\loci$. We then pass $x_{i}$ through a convolutional architecture, the embedding space of which can be represented as $
      z_{i}^{c} = f^{CNN}(x_{i})
$, 
where $f^{CNN}$ is parameterized by a deep convolutional network as shown in Fig.~\ref{fig:workflow} and returns $z_{i}^{c} \in \mathbb{R}^{512}$. We then pass $z_{i}^{c}$ through a nonlinear function consisting of two dense layers with ReLU activations obtaining a vector in $\mathbb{R}^{256}$. 
Next, we concatenate representations from the two different modalities 
and use another 
four dense layers, each with a ReLU activation, and a final regression layer to produce the final estimate.
The proposed architecture, as shown in Fig.~\ref{fig:workflow}, is jointly trained on the task of poverty prediction. However, we freeze the weights of the $Doc2Vec$ during training on the poverty task.


\input{00_workflow.tex}



%% file: 00_article_distances.tex
\begin{figure}
    \subfloat[]{
        \centering
        \includegraphics[width=3.35in]{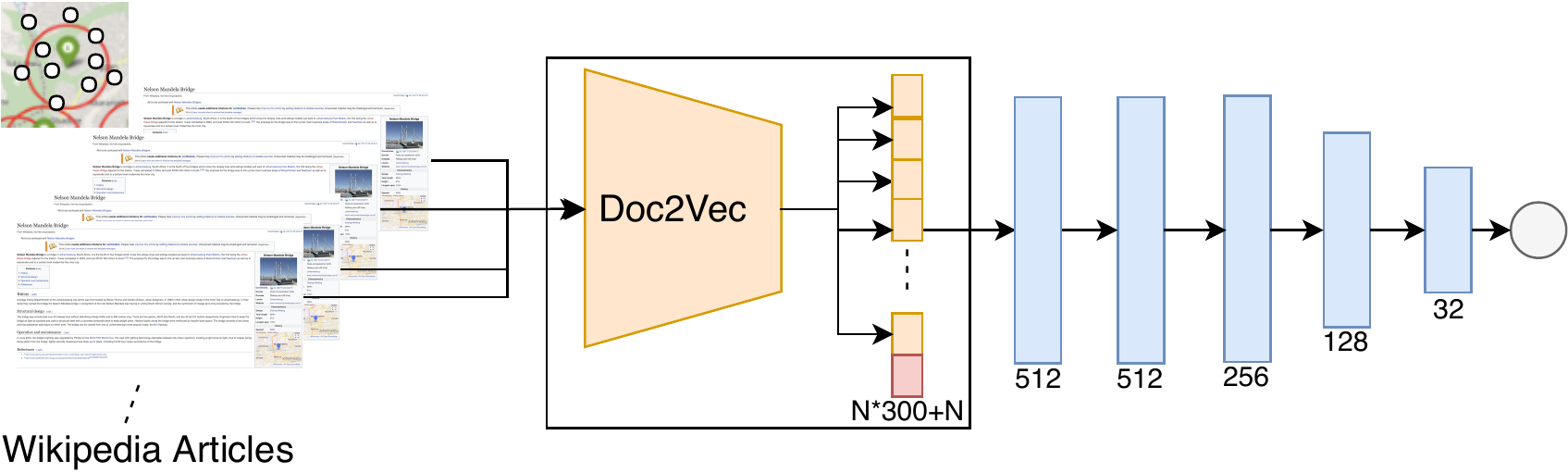}
        \label{fig:wikipedia_only_model}
    }
    
    \subfloat[]{
            \centering
            \includegraphics[width=1.6in]{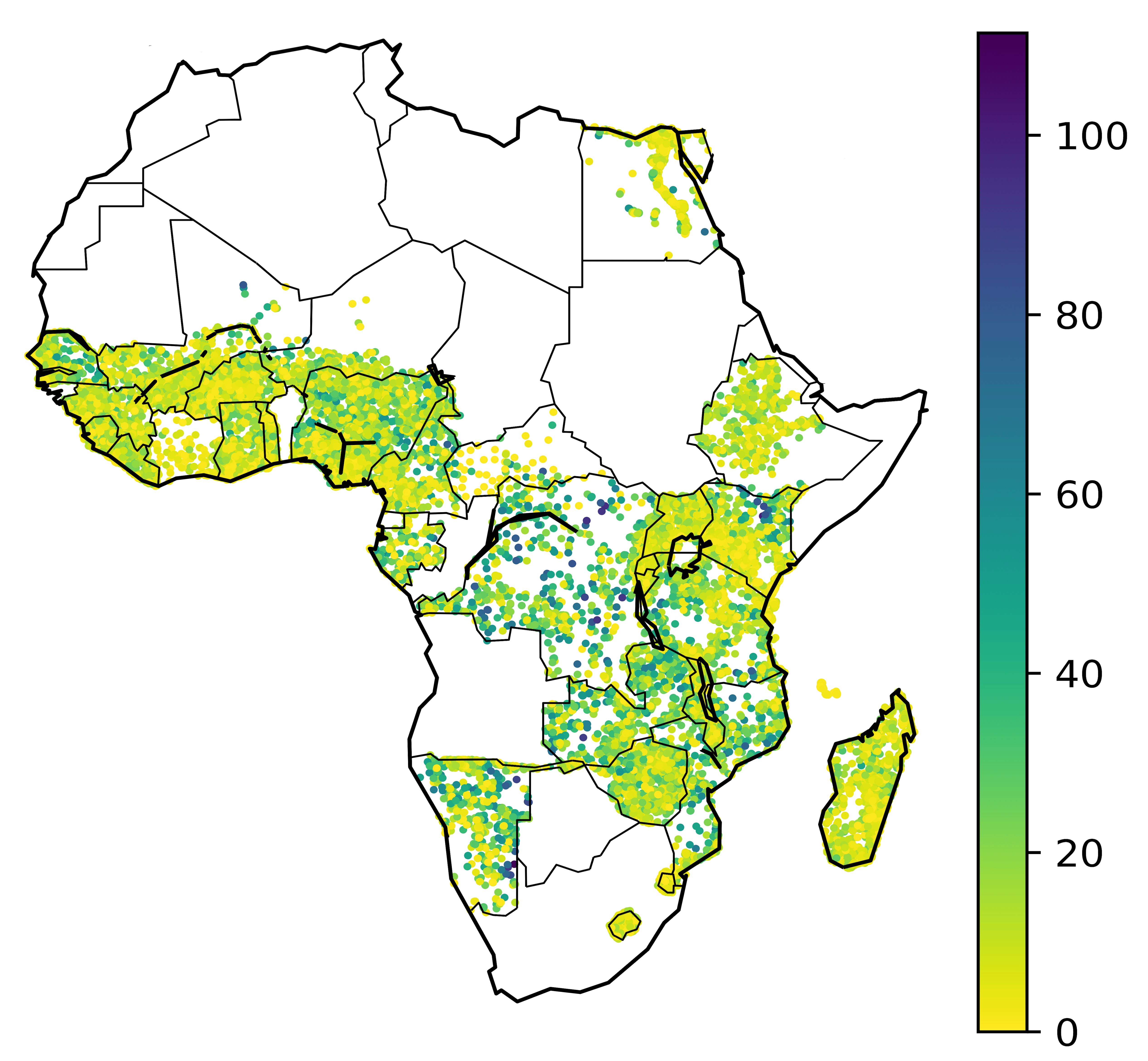}
            \label{fig:article_distances_1}
    }
    \subfloat[]{
            \centering
            \includegraphics[width=1.6in]{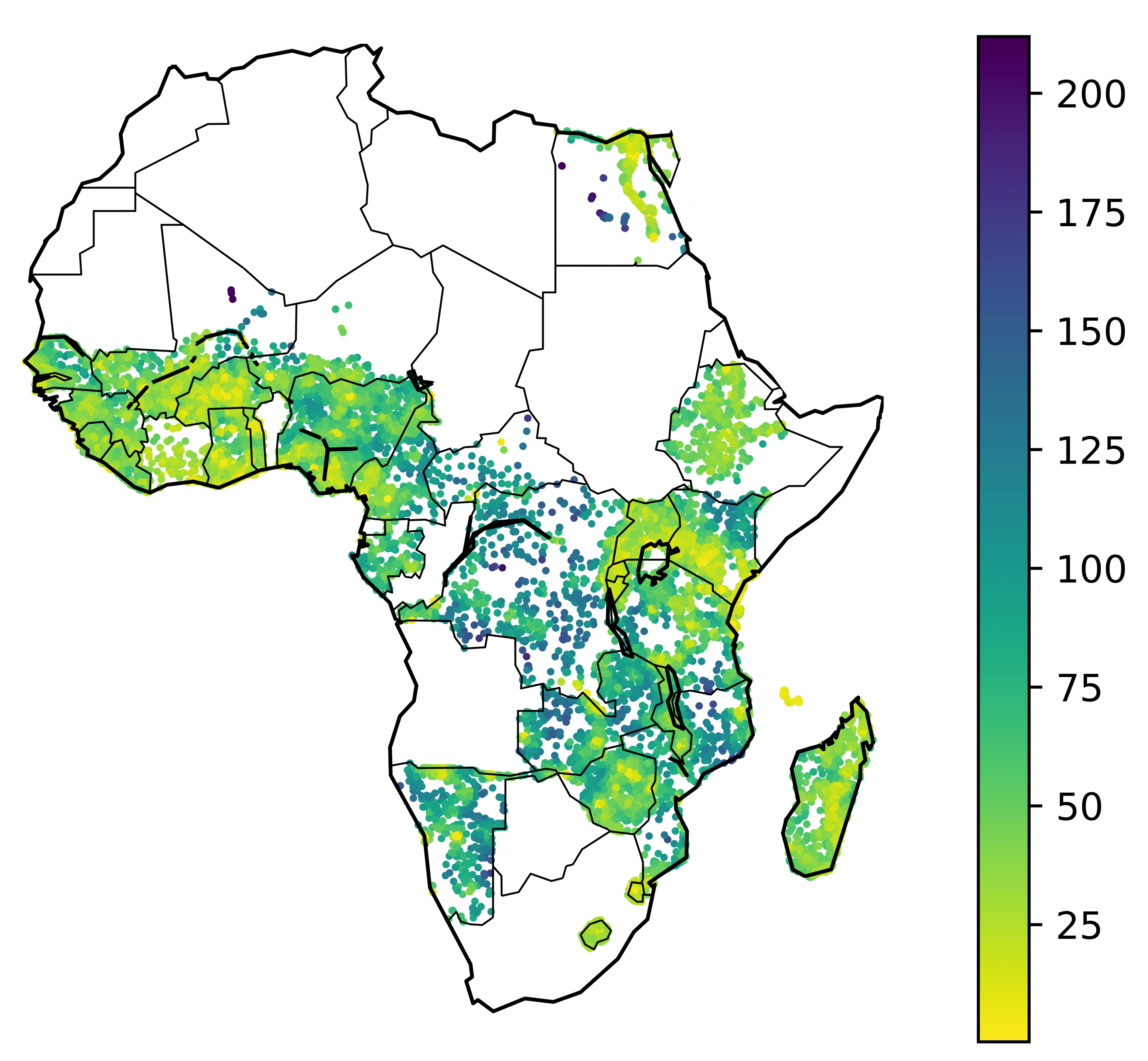}
            \label{fig:article_distances_10}
    }
    \caption{\textbf{Top: } The proposed Wikipedia Embedding Model is shown in (a). Our architecture uses textual information to regress on economic development. For a given DHS location, we use the $N$=$10$ closest Wikipedia articles. The blue part is trained on the DHS prediction task, whereas the encoder part of the \textit{Doc2Vec} is pre-trained on the global geolocated Wikipedia article corpus and fixed during training. Panels (b) and (c) show the average distance (km) to the closest and 10th closest Wikipedia articles for each survey location.
    }
    \label{Article Distances}
\end{figure}

%% file: 00_workflow.tex
\begin{figure}
\centering
\includegraphics[width=3.35in]{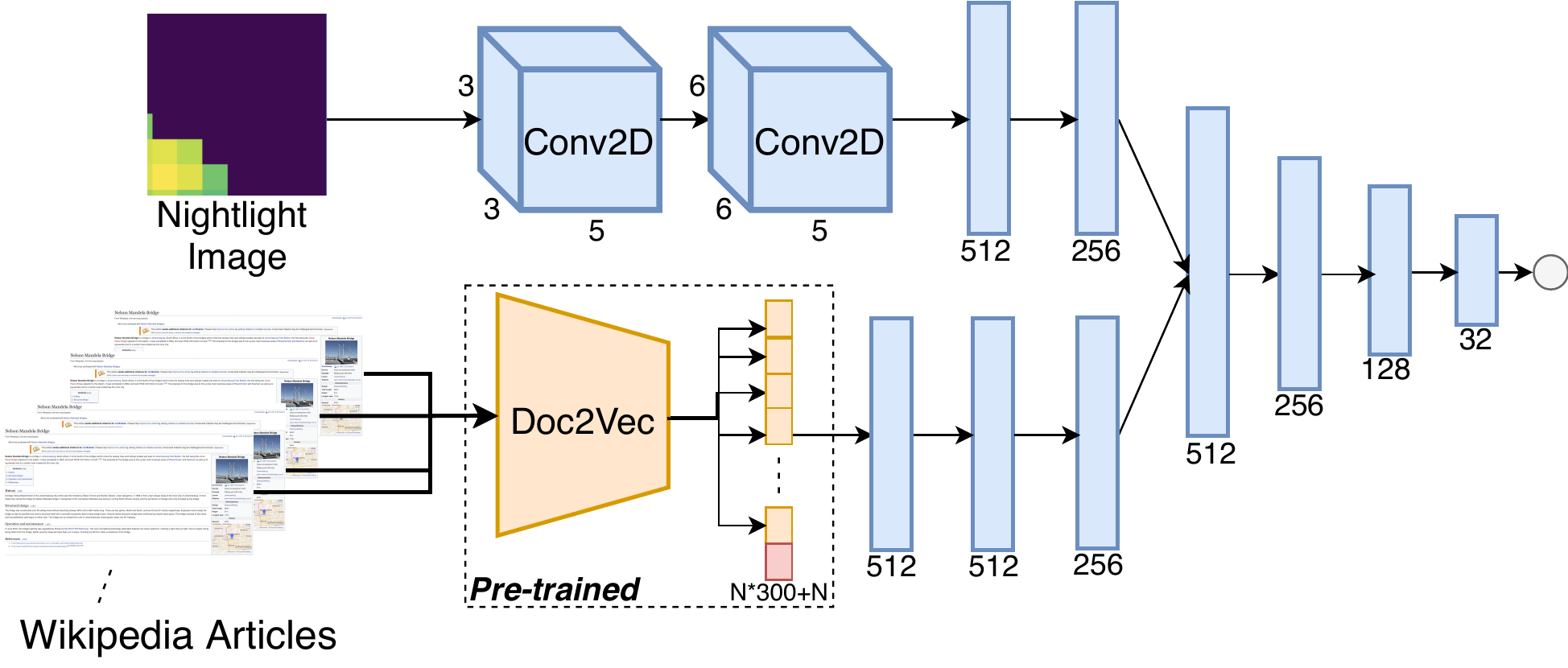}
\caption{The proposed Multi-Modal architecture. Our architecture uses visual and textual information to regress on survey data. For a given location, we use nightlight imagery and the Wikipedia articles within a certain neighborhood. The blue part is jointly trained on the survey prediction task, whereas the encoder part of \textit{Doc2Vec} is pre-trained on the global geolocated Wikipedia article corpus and kept frozen. 
}
\label{fig:workflow}
\end{figure}

%% file: 0_results_poverty.tex
\section{Poverty Prediction}

\input{00_table_poverty.tex}

Our study proposes novel single and multi-modal deep neural networks that predict economic development using a combination of Wikipedia and remotely sensed data (nightlights). To compare with prior work and put our results into perspective, we also consider a baseline \nightonly{} model,
which only uses nightlights as an input. Our \nightonly{} model is based on the ``nightlights branch" of the proposed multi-modal architecture in Fig.~\ref{fig:workflow}. This allows us to directly evaluate the contribution of Wikipedia articles in the multi-modal model. We emphasize that this is a strong baseline model, as previous works use simpler, hand-crafted features, such as the mean intensity or simple summary statistics (max, min, median), instead of the image \cite{noor2008using}. Such simpler nightlight models are not included in this study as they significantly underperform our CNN-based \nightonly{}  model. We additionally compare to a model from~\cite{Jean790} that uses a combination of nighlights and high-resolution satellite images.

All the architectures are evaluated on three experiments:
\begin{itemize}
    \item [(a)] Intra-nation tests: Train and test on the same country.
    \label{intra}
    \item [(b)] Cross-national boundary testing between African countries: Train on one country and test on another.
    \item [(c)] Leave-one-country-out training: Train on all countries except one and test on that country, and vice versa. 
\end{itemize}
These experiments were chosen as they reflect the possible real-world applications of this model. 
In particular, (a) intra-nation inference can be used to fill in gaps of surveys within a country. Data acquired using this method can be very useful in providing local government with the tools to efficiently target aid. 
Cross-national boundary (b) and leave-one-out based predictions (c), on the other hand, can be used to make predictions in countries with no recent survey data. 

As we regress directly on the poverty index, we measure the accuracy of our model using the square of the Pearson correlation coefficient (Pearson's $r^2$). 
This metric was chosen so that comparative analysis could be performed with previous literature \cite{Jean790,nips_2017_workshop_stefano}. Further, in cases of out-of-distribution testing, such as cross-national boundary tests, linear correlation can still be a meaningful metric. For example, our model may have slightly lower predictions if it is trained on a relatively poor country and tested on a wealthy one.
In this case, comparing the absolute predicted value with the ground truth would not be a fair option. On the other hand, Pearson's $r^2$ is invariant under separate changes in location and scale between the two variables. This allows the metric to still provide insight into the ability of the model to distinguish between poverty levels. In the space of aid, knowing relative wealth indices is often a good alternative to knowing the true ground truth values, as it still provides insight on where to target aid. 


\subsection{Wikipedia Embedding Model}

We train our \wikipedia{} (WE) model using a Mean Squared Error loss function, along with an Adam optimizer with a learning rate of 0.001. We report in Table \ref{table:Results Table} $r^2$ values obtained from  (a) intra-nation (shaded in blue), as well as (b) cross-country  and (c) leave-one-out evaluations. 

\textbf{In-Country:} We observe that our \wikipedia{} (WE) model is able to robustly learn associations with poverty. Compared with the \nightonly{} (NL) model, we see that WE outperforms NL in all 5 countries. Comparing to the state of the art results from~\cite{Jean790}, which are based on a combination of nighlights and high-resolution daytime imagery, we see that we are consistently within $10$-$15\%$ of the reported $r^2$'s for training and testing on the same countries, except for Tanzania, where we improve from .58 to .64. 

\textbf{Out-of-Country:} 
Cross national boundary experiments allow us to test the generalization capabilities of the model. As we are taking 5 km $\times$ 5 km nightlight images of a data point, there may be portions of images similar to both train and test sets because of underlying spatial correlations between the examples. This can also happen for the Wikipedia embeddings, as the $N$ = $10$ closest articles can have some overlap between training and testing examples when evaluating within the same country. Prediction across national boundaries is a more difficult task, both because of covariate shift and because there is almost no spatial correlation between the train and test sets the model can exploit. 

\begin{figure}[h!]
    \centering
    \includegraphics[width=3.5in]{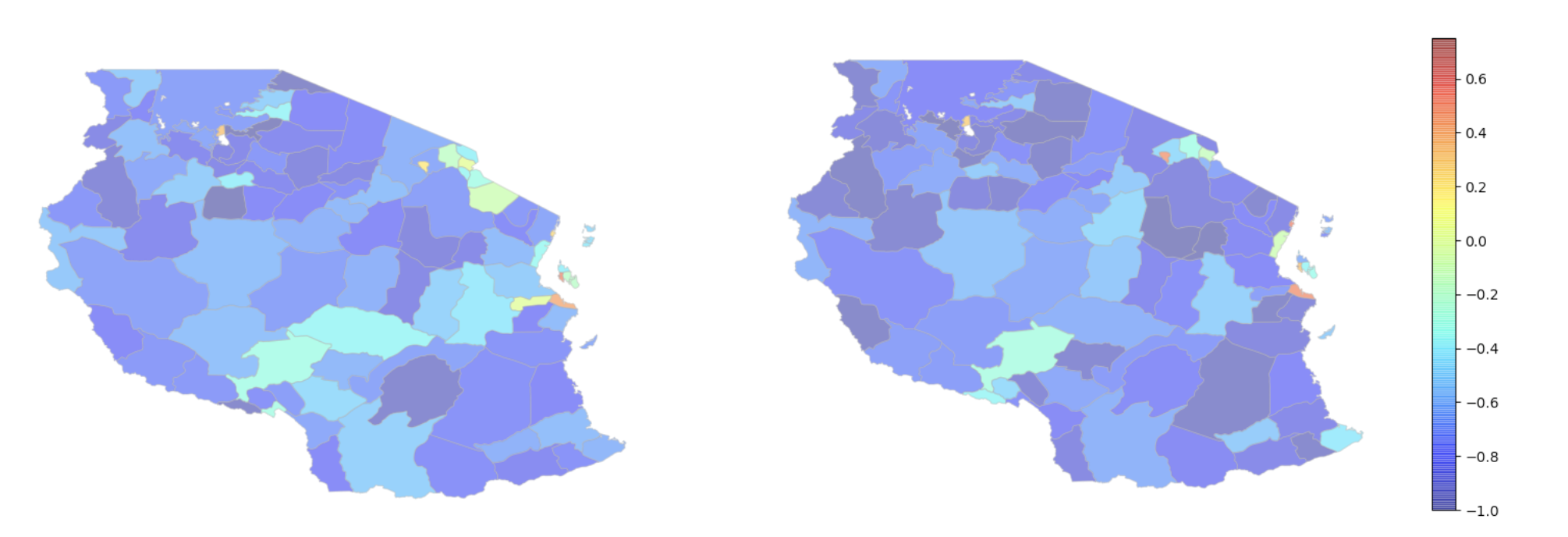}
    \caption{\textbf{Left:} Ground truth values of 1074 data points averaged on Admin 2 Level for Tanzania. \textbf{Right:} Wikipedia Embedding Model predicted values averaged on Admin 2 level for a model trained on Ghana and tested on Tanzania ($r^2=0.52$). The resulting predictions show that the model is capable of generalizing across national boundaries.}
    \label{Combined Raster}
\end{figure}

The details of these results are in Table~\ref{table:Results Table}, where the statistics of a model trained on country A and evaluated on country B are located in the column labelled A and the row labelled B. 
The ``Average'' row simply presents the average performance of the 
models for each column country.
For example, Fig.~\ref{Combined Raster} displays the results of the WE model when it is trained on the Ghanaian article distribution and then evaluated on the Tanzanian article distribution. We observe that, while the model may not always predict the correct wealth level for a region, it is often correct in its \emph{relative} predictions for adjacent regions and regions within the nation in general. This suggests that scaling the model by the mean income level for a country (data which is available) may allow the model to correctly determine both relative and absolute wealth levels.

Comparing WE to NL for out-of-country experiments, we see that NL often outperforms WE by a very slight margin, usually by approximately .1 to .15 in terms of $r^2$. This is also true when comparing WE to the out-of-country state of the art results from \cite{Jean790}. The exception to both these trends is that WE outperforms both NL and \cite{Jean790} by about .05 to .1 on over half the models either trained or tested on Malawi. Throughout all these results, the core theme observed is that the WE model is learning generalizable features and not simply picking up on spatial correlations, since the train and test sets are \emph{entirely} disjoint.

To better understand the role of Wikipedia embeddings, we 
visualize the predictions of the NL and WE models trained on Ghana and tested on Tanzania in Fig.~\ref{Prediction}. Because certain countries' nightlights datasets are dominated by dark images, in certain circumstances the \nightonly{} model is prone to predict constant values, which results in the horizontal line of predictions shown in the left plot of Fig.~\ref{Prediction}. However, the \wikipedia{} model does not suffer from such a shortcoming and is able to increase the range of the predictions, 
as shown in the right plot of Fig.~\ref{Prediction}.


\input{00_nl_we.tex}

\textbf{Leave-One-Out:} 
The column ``All'' in Table~\ref{table:Results Table} represents models trained using the leave-one-out approach, such that a model in column ``All'' evaluated on nation $B$ was trained on all countries but $B$.
The inverse is true for row ``All'', where a model trained on nation A is
tested on all other countries.

Examining the leave-one-out experiments, we observe that training with this approach achieves the best results on average compared to training on any other single country. The average $r^2$ of WE models trained in this way is .49, the highest such WE model averages. Interestingly, the models trained on only Tanzania perform similarly or even better than the ones trained on All countries on the Malawi test samples. We hypothesize this is due to similar patterns of the neighboring countries in the dataset.

\subsection{Multi-Modal Model}

We train our \multimodal{} (MM) model using a Mean Squared Error loss function, along with an Adam optimizer with a learning rate of 0.001.
Table \ref{table:Results Table} displays the Pearson's $r^2$ results of the model. Comparing WE with MM, we observe that in general the accuracy ($r^2$) of the models 
for out-of-country evaluation 
improves by almost 50\% across the board for the multi-modal approach. Importantly, the proposed multi-modal architecture outperforms the other models by large margin when trained via the leave-one-out approach, proving that the textual branch maintains its contribution even when training the CNN with large amounts of nightlights data. Some models, such as the one trained on Uganda and tested on Nigeria, obtain over $100\%$ improvement, while others, such as the one trained on Ghana and tested on Tanzania, obtain less than a $20\%$ improvement. Importantly, all train-test country combinations see some degree of improvement. When considering the results of our ablative \nightonly{} model, we also observe that the MM achieves superior results for both intra-country and cross-boundary tests nearly across the board. 
Such results suggest that Wikipedia embeddings and nightlight images provide highly complementary information about poverty. Finally, the MM outperforms state of the art by an average of $31.2\%$ \cite{Jean790}, with improvements nearly across the board. 

%% file: 00_table_poverty.tex
\begin{table*}[t]
\scriptsize
\centering
\begin{tabular}{l c c c c c c c c c c c c c c c c c c}
\toprule
& \multicolumn{18}{c}{Trained on} \\
\cmidrule(lr){2-19}
& \multicolumn{3}{c}{Ghana} & \multicolumn{3}{c}{Malawi} & \multicolumn{3}{c}{Nigeria} & \multicolumn{3}{c}{Tanzania} & \multicolumn{3}{c}{Uganda} & \multicolumn{3}{c}{All} \\
\cmidrule(lr){2-4}
\cmidrule(lr){5-7}
\cmidrule(lr){8-10}
\cmidrule(lr){11-13}
\cmidrule(lr){14-16}
\cmidrule(lr){17-19}
Tested on & NL & WE & MM & NL & WE & MM & NL & WE & MM & NL & WE & MM & NL & WE & MM & NL & WE & MM \\
\midrule
Ghana & \cellcolor{blue!25} 0.41 & \cellcolor{blue!25} 0.47 & \cellcolor{blue!25} \textbf{0.76} &0.43 &0.42 &\textbf{0.61} &\textbf{0.64} &0.37 &0.45 &0.46 &0.44 &\textbf{0.51} &\textbf{0.65} &0.34 &0.58 &\textbf{0.61} &0.40 & 0.60\\
Malawi &0.30 &0.40 &\textbf{0.48} & \cellcolor{blue!25} 0.24 & \cellcolor{blue!25} 0.49 & \cellcolor{blue!25} \textbf{0.64} &0.34 &0.35 &\textbf{0.55} &0.37 &0.42 &\textbf{0.56} &0.34 &0.25 &\textbf{0.52} &0.40 &0.38 &\textbf{0.56} \\
Nigeria &0.44 &0.32 &\textbf{0.60} &0.31 &0.37 &\textbf{0.52} & \cellcolor{blue!25} 0.30 & \cellcolor{blue!25} 0.52 & \cellcolor{blue!25} \textbf{0.70} &0.46 &0.37 &\textbf{0.57} &0.48 &0.24 &\textbf{0.57} &0.48 &0.35 &\textbf{0.61}\\
Tanzania &0.50 &0.52 &\textbf{0.58} &0.46 &0.52 &\textbf{0.63} &0.52 &0.48 &\textbf{0.64} & \cellcolor{blue!25} 0.60 & \cellcolor{blue!25} 0.64  & \cellcolor{blue!25} \textbf{0.71} &0.52 &0.49 &\textbf{0.63} &0.54 &0.50 &\textbf{0.59}\\
Uganda &0.61 &0.45 &\textbf{0.70} &0.58 &0.50 &\textbf{0.74} &0.62 &0.40 &\textbf{0.70} &0.64 &0.49 &\textbf{0.75} & \cellcolor{blue!25} 0.53 & \cellcolor{blue!25} 0.57 & \cellcolor{blue!25} \textbf{0.76} &0.62 &0.52 &\textbf{0.71} \\
All &0.44 &0.32 &\textbf{0.46} &\textbf{0.55} &0.26 &0.51 &\textbf{0.51} &0.37 &0.48 &0.49 &0.32 &\textbf{0.65} &0.46 &0.27 &\textbf{0.48} & \cellcolor{blue!25} 0.45 & \cellcolor{blue!25} \textbf{0.77} & \cellcolor{blue!25} 0.76 \\
\midrule
Average &0.45 &0.41 &\textbf{0.60} &0.43 &0.43 &\textbf{0.61} &0.49 &0.42 &\textbf{0.59} &0.50 &0.45 &\textbf{0.63} &0.50 &0.36 &\textbf{0.59} &0.52 &0.49 &\textbf{0.64} \\
\bottomrule
\label{table:Results Table}
\end{tabular}
\caption{Performances of Nightlight-Only (NL), Wikipedia Embedding (WE), and Multi-Modal (MM) models on in-country and out-of-country experiments. Columns and rows represent the countries the models were trained and tested on, respectively. The Multi-Modal model outperforms the other models on both in-country (shaded) and cross-country experiments.
}
\label{table:Results Table}
\end{table*}

%% file: 00_nl_we.tex
\begin{figure}[h]
   \subfloat{
        \centering
        \includegraphics[width=1.6in]{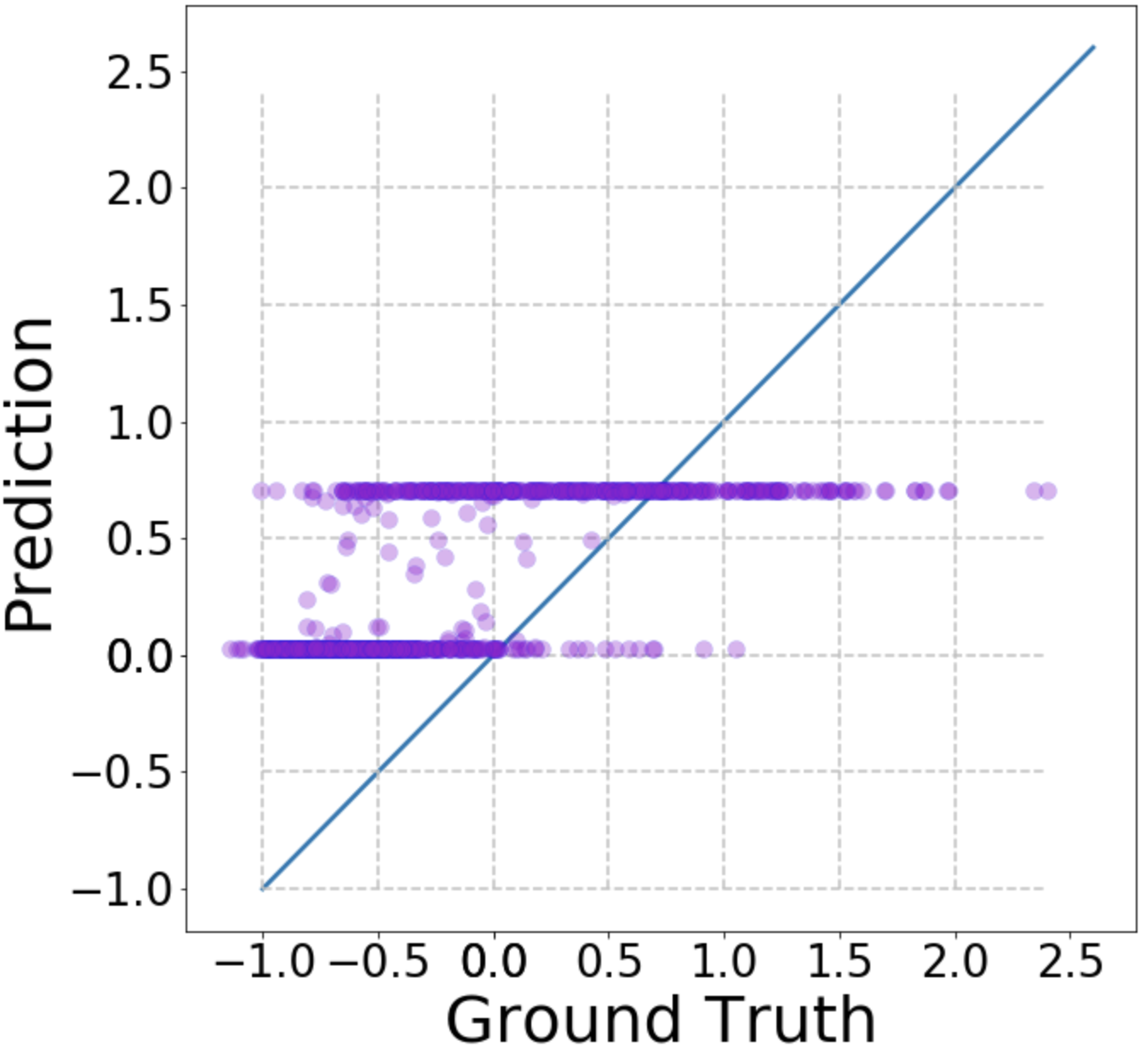}
   }
   \subfloat{
        \centering
        \includegraphics[width=1.6in]{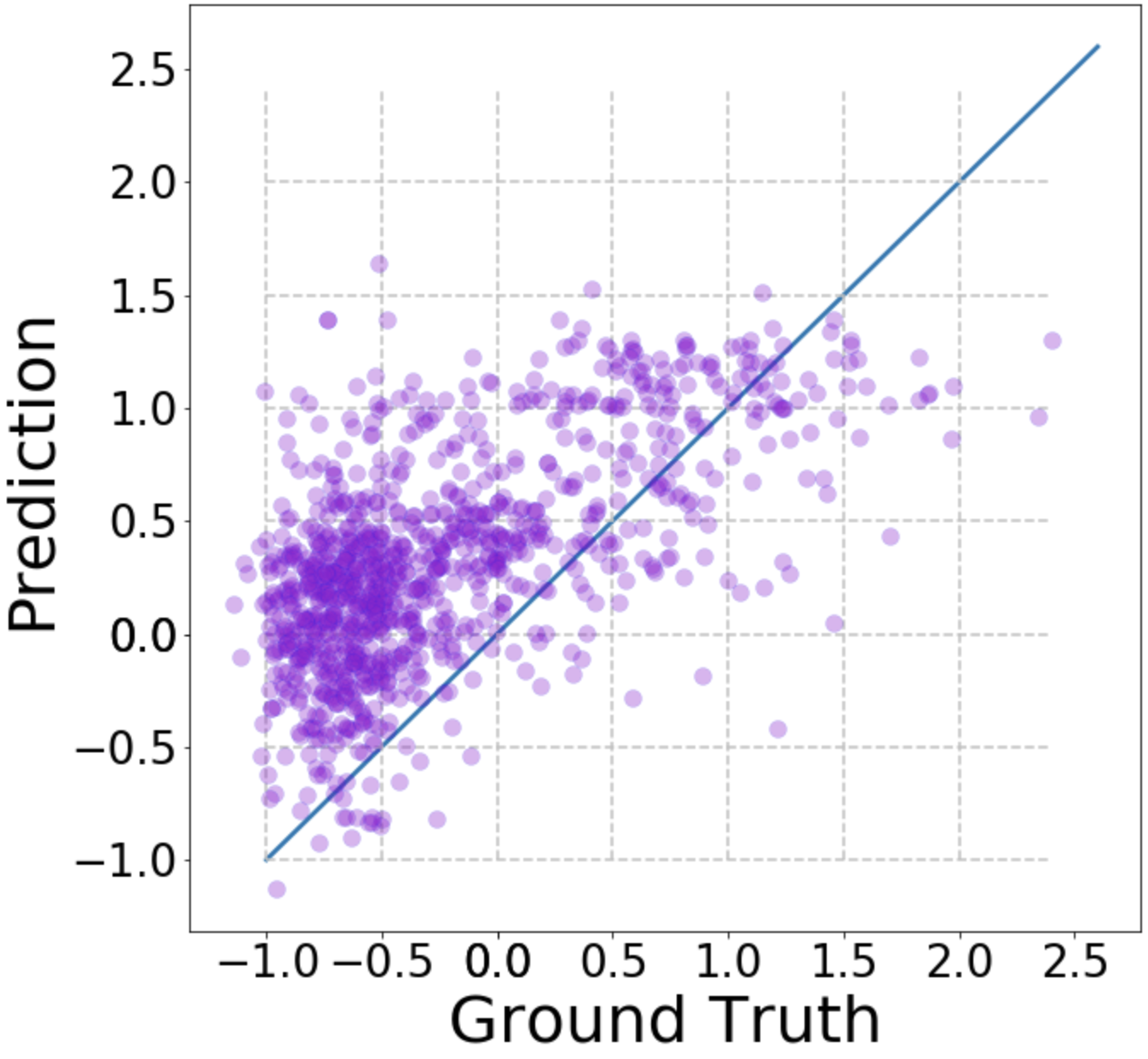}
        }
    \caption{
    Wealth prediction for Nightlight-Only 
    (\textbf{Left})  
    and Wikipedia Embedding (\textbf{Right}) models trained on Ghana and tested on Tanzania.  Nightlights-Only model produces highly correlated results for different ground points, resulting in horizontal lines. Meanwhile, the Wikipedia Embeddings model leads to a better range of predictions.
    }
    \label{Prediction}
\end{figure}

%% file: 0_interpretability.tex
\subsection{Ablation study and Interpretability}



So far we have presented our results for both architectures using $N=10$ closest articles to the ground truth point. To evaluate the sensitivity to $N$, we conducted additional experiments by varying number of articles $N$. As shown in Fig.~\ref{Article Iteration Graph}, considering average performance, we can conclude that $N=10$ provides good performance for most of the countries.
However, for some countries (Uganda, Tanzania), even a few articles can provide optimal models. We believe this might be due to the different density of articles (see Fig.~\ref{fig:article_distances_1}, and ~\ref{fig:article_distances_10}) around data points. 

\input{00_number_articles.tex}
\input{00_pca_analysis.tex}

To help us better understand Wikipedia article embeddings, we performed a PCA analysis of the learned Wikipedia article embeddings. More specifically, we trained a model on Tanzania, evaluated it on Uganda, ranked all its projections from richest to poorest, and then stored the top third and bottom third of these points, corresponding to the richest $33\%$ and the poorest $33\%$ of all projections for Uganda. For each of the points in these wealthiest and poorest thirds, we averaged the embeddings of the 10 closest articles to it. We then compared these averaged embeddings' PCA projections to the projections of `school', `university', `hospital', `company', and `settlement' articles that appeared in the test set (see Fig.~\ref{pca analysis}). We observed that the average \textit{Doc2Vec} embeddings of the wealthier coordinates clustered around article categories which intuitively are related to wealth. In particular, wealthier places tended to cluster toward `school', `university', `hospital', and `company' embeddings, while poorer places were more related to `settlement' articles.

%% file: 00_number_articles.tex
\begin{figure}
    \centering
    \includegraphics[width=0.350\textwidth]{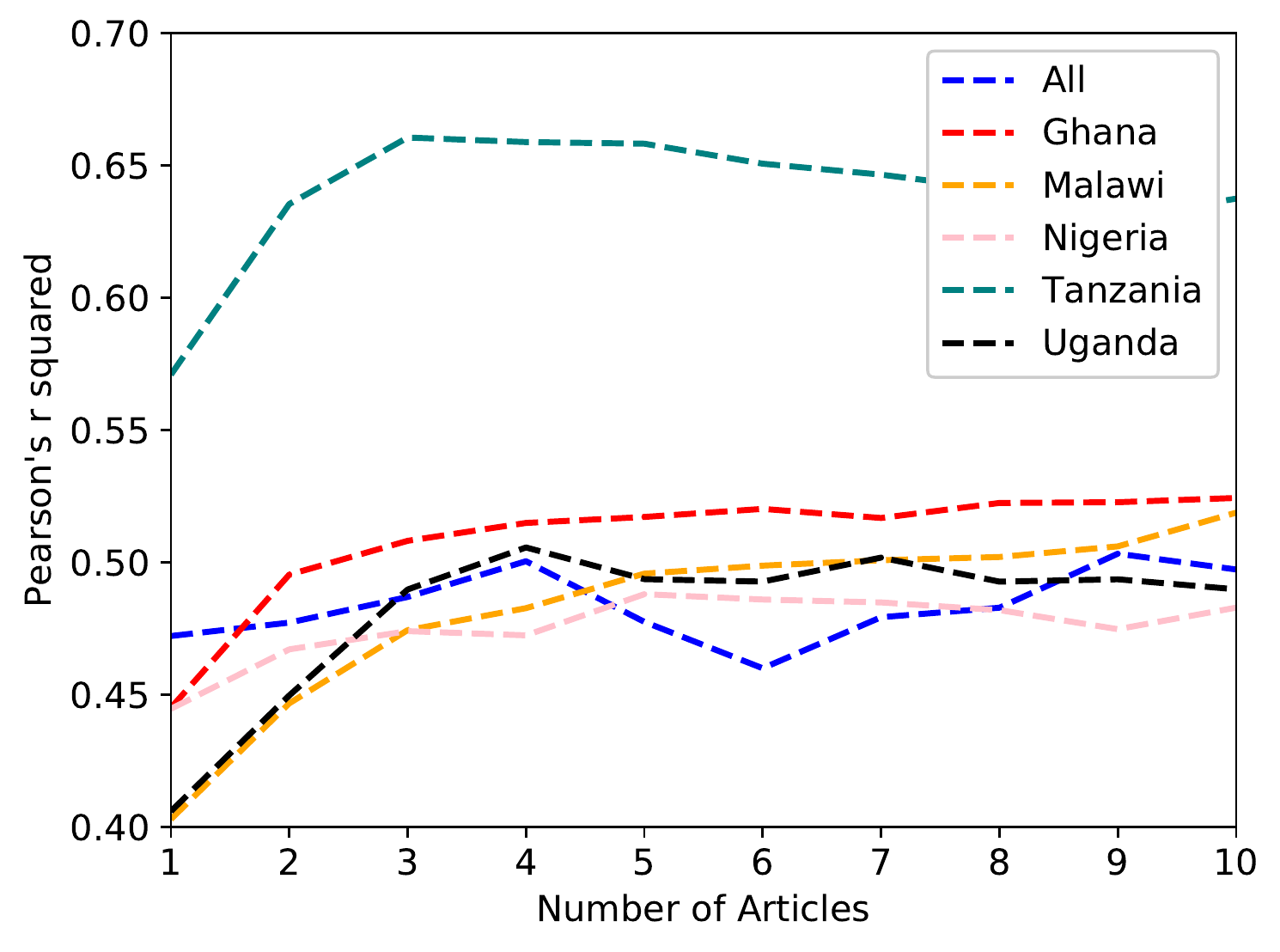}
    \caption{Pearson's $r^2$ performance for models trained using varying $N$ geographically nearest article embeddings, trained on the specified country and evaluated on Tanzania. 
    }
    \label{Article Iteration Graph}
\end{figure}

%% file: 00_pca_analysis.tex
\begin{figure}[h]
\centering
\includegraphics[width=2in]{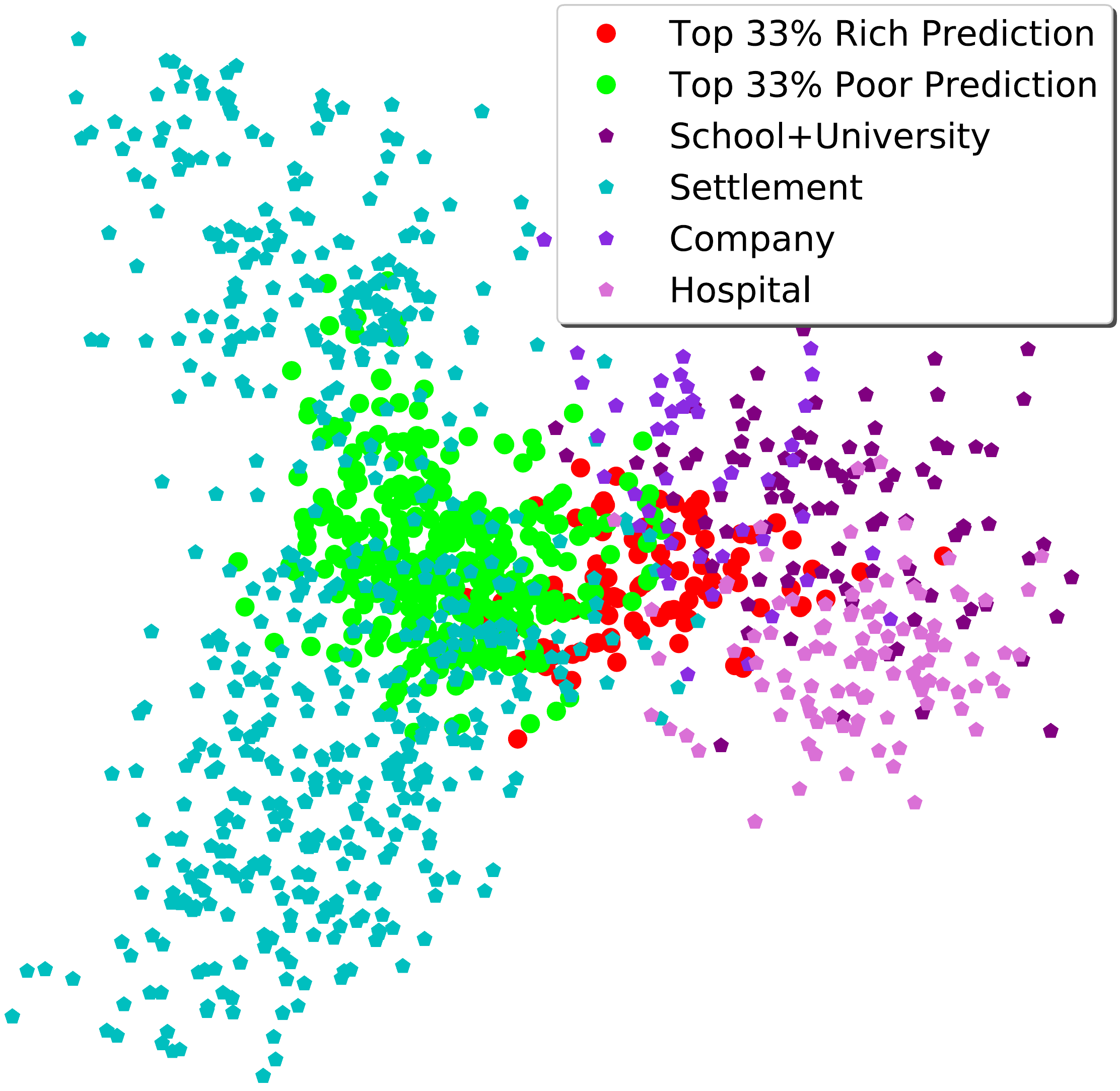}
\caption{Wikipedia Embedding PCA Analysis. 
The WE model learns to  project the wealthiest $33\%$ quantile to the space closer to \textit{Company}, \textit{Hospital} and \textit{School+University} related Wikipedia Articles' embeddings.} 
\label{pca analysis}
\end{figure}

%% file: 0_results_education.tex
\section{Education Index Prediction}


To further test the effectiveness of our text-only model in predicting other socioeconomic outcomes, we train and evaluate our model on a new task: education level-prediction.
Improving the quality of global education is the 4th SDG --- access to education is foundational to the creation of self-sustaining and self-improving societies.

In addition to predicting a different type of outcome, we use a new source of ground truth data for this task. We use the Afrobarometer R6 Codebook dataset \cite{AFRO}, which is distributed by Afrobarometer, a research network that conducts public attitude surveys on democracy, governance, economic conditions, and related issues in more than 35 countries in Africa. We utilize the `eaeducationlevel' index, which has values ranging from 1 to 4, with larger number standing for higher level of education (i.e., 1 stands for `No formal schooling', 4 stands for `post-secondary qualifications'), for our groundtruth labels. To be consistent with the previous experiments on poverty, we limit our research to 5 countries (see Table \ref{table:Education Results Table}) in Africa, each with 300 datapoints. Notice that Afrobarometer is sampled at different locations compared to DHS.
We evaluate our Wikipedia Embedding model on the (a) Intra-nation and (b) Cross-Boundary training tasks, as in the previous section. We use the same training procedure, except we regress on a different outcome.

Table \ref{table:Education Results Table} displays our results on this task. We notice that that in-country models (shaded) consistently outperform out-of-country models, as was generally the case for the poverty task. However, on average the performance is worse compared to the poverty task, with averages ranging from .20-.25 compared to .40-.45. One possible explanation is that education is a finer-grained indicator that is more difficult to predict from Wikipedia. We can expect improved results with the inclusion of more education-related articles in future releases.
Nevertheless, this experiment suggests that Wikipedia provides useful information to predict multiple socioeconomic indicators, across different countries and even different survey methodologies.



\input{00_table_education.tex}

%% file: 00_table_education.tex
\begin{table}[h]
\scriptsize
\centering
\begin{tabular}{c c c c c c c c c c c c c c c c c c}
\toprule
& \multicolumn{4}{c}{Trained on} \\
\cmidrule(lr){1-8}
{Tested on} &{Ghana} & {Malawi} &{Nigeria} & {Tanzania} &{Uganda}\\
\cmidrule(lr){1-3}
\cmidrule(lr){3-4}
\cmidrule(lr){4-5}
\cmidrule(lr){5-6}
\cmidrule(lr){6-7}
\cmidrule(lr){7-8}
Ghana &\cellcolor{blue!25} \textbf{0.31} &0.09 &0.11 &0.22 &0.28\\
Malawi &0.21 &\cellcolor{blue!25} \textbf{0.32}  &0.09 &0.23 &0.24\\
Nigeria &0.32 &0.11 &\cellcolor{blue!25} \textbf{0.39} &0.24 &0.23\\
Tanzania &0.27 &0.20 &0.10 &\cellcolor{blue!25} \textbf{0.34} & 0.30\\
Uganda &0.29 &0.19 &0.10 &0.25 &\cellcolor{blue!25} \textbf{0.30}\\
\midrule
Average &0.28 &0.18 &0.16 &0.26 &0.27\\
\bottomrule
\label{table:Education Results Table}
\end{tabular}
\caption{Performances of Wikipedia Embedding model on education level prediction as measured by Pearson's $r^2$. Bolding indicates the best model in each country.}
\label{table:Education Results Table}
\end{table}

%% file: 0_conclusion.tex
\section{Conclusion}

In this paper, we propose a novel method for the utilization of geolocated Wikipedia articles for socioeconomic applications. We do this by obtaining vector representations of articles via the $Doc2Vec$ NLP model. We then pair these latent embeddings with survey data and evaluate models to predict poverty and education outcomes: a \wikipedia{} model, and a \multimodal{} model that also uses nightlights information. Using this framework, we find Wikipedia articles to be informative about socioeconomic indicators. To the best of our knowledge, this is the first time Wikipedia is used in this way. Additionally, the \multimodal{} model outperforms current state-of-the-art results for the poverty prediction task by over $20\%$. 
Our findings suggest this pipeline, and the geolocated Wikipedia article dataset, has applications not just in poverty analysis, but also more general socioeconomic prediction, such as education and health related outcomes. This novel approach could lead to advances in development economics, where studies often rely only on nightlights information. Additionally, we hope this approach will accelerate progress towards the UN SDGs by improving the way we estimate lacking socioeconomic indicators, particularly in developing countries, with the aim of improving responses from regional governments and international aid organizations.\\




\noindent\textbf{Reproducibility:} Upon publication, we will publicly release our Wikipedia, nightlights and processed indexes data. Raw DHS surveys can be obtained from the World Bank, while Afrobarometer is available through AidData.

%% file: 0_main.bbl
\begin{thebibliography}{}

\bibitem[\protect\citeauthoryear{AFR}{}]{AFRO}
The afro program website.
\newblock \url{http://http://afrobarometer.org}.

\bibitem[\protect\citeauthoryear{Antwi-Agyei \bgroup \em et al.\egroup
  }{2012}]{antwi2012mapping}
Philip Antwi-Agyei, Evan~DG Fraser, Andrew~J Dougill, Lindsay~C Stringer, and
  Elisabeth Simelton.
\newblock Mapping the vulnerability of crop production to drought in ghana
  using rainfall, yield and socioeconomic data.
\newblock {\em Applied Geography}, 32(2):324--334, 2012.

\bibitem[\protect\citeauthoryear{Blumenstock \bgroup \em et al.\egroup
  }{2015}]{blumenstock2015predicting}
Joshua Blumenstock, Gabriel Cadamuro, and Robert On.
\newblock Predicting poverty and wealth from mobile phone metadata.
\newblock {\em Science}, 350(6264):1073--1076, 2015.

\bibitem[\protect\citeauthoryear{Carneiro and
  Mylonakis}{2009}]{carneiro2009google}
Herman~Anthony Carneiro and Eleftherios Mylonakis.
\newblock Google trends: a web-based tool for real-time surveillance of disease
  outbreaks.
\newblock {\em Clinical infectious diseases}, 49(10):1557--1564, 2009.

\bibitem[\protect\citeauthoryear{Dang and Ignat}{2016}]{dang2016quality}
Quang~Vinh Dang and Claudia-Lavinia Ignat.
\newblock Quality assessment of wikipedia articles without feature engineering.
\newblock In {\em Proc. of the 16th Joint Conference on Digital Libraries},
  pages 27--30. ACM, 2016.

\bibitem[\protect\citeauthoryear{Dang and Ignat}{2017}]{dang2017end}
Quang-Vinh Dang and Claudia-Lavinia Ignat.
\newblock An end-to-end learning solution for assessing the quality of
  wikipedia articles.
\newblock In {\em Proc. of the 13th International Symposium on Open
  Collaboration}, page~4. ACM, 2017.

\bibitem[\protect\citeauthoryear{DHS}{}]{DHS}
The dhs program website.
\newblock \url{http://www.dhsprogram.com}.

\bibitem[\protect\citeauthoryear{Elvidge \bgroup \em et al.\egroup
  }{2017}]{VIIRS}
Christopher~D Elvidge, Kimberly Baugh, Mikhail Zhizhin, Feng~Chi Hsu, and
  Tilottama Ghosh.
\newblock Viirs night-time lights.
\newblock {\em Int. J. Remote Sens.}, 38(21):5860--5879, November 2017.

\bibitem[\protect\citeauthoryear{Filmer and Pritchett}{1999}]{filmer1999effect}
Deon Filmer and Lant Pritchett.
\newblock The effect of household wealth on educational attainment: evidence
  from 35 countries.
\newblock {\em Population and development review}, 25(1):85--120, 1999.

\bibitem[\protect\citeauthoryear{Haklay and
  Weber}{2008}]{haklay2008openstreetmap}
Mordechai Haklay and Patrick Weber.
\newblock Openstreetmap: User-generated street maps.
\newblock {\em Ieee Pervas Comput}, 7(4):12--18, 2008.

\bibitem[\protect\citeauthoryear{Jean \bgroup \em et al.\egroup
  }{2016}]{Jean790}
Neal Jean, Marshall Burke, Michael Xie, W.~Matthew Davis, David~B. Lobell, and
  Stefano Ermon.
\newblock Combining satellite imagery and machine learning to predict poverty.
\newblock {\em Science}, 353(6301):790--794, 2016.

\bibitem[\protect\citeauthoryear{Kahn \bgroup \em et al.\egroup
  }{2007}]{kahn2007research}
Kathleen Kahn, Stephen~M Tollman, Mark~A Collinson, Samuel~J Clark, Rhian
  Twine, Benjamin~D Clark, Mildred Shabangu, Francesc~Xavier Gomez-Olive, Obed
  Mokoena, and Michel~L Garenne.
\newblock Research into health, population and social transitions in rural
  south africa: Data and methods of the agincourt health and demographic
  surveillance system1.
\newblock {\em Scandinavian Journal of Public Health}, 35(69\_suppl):8--20,
  2007.

\bibitem[\protect\citeauthoryear{Lau and Baldwin}{2016}]{lau2016empirical}
Jey~Han Lau and Timothy Baldwin.
\newblock An empirical evaluation of doc2vec with practical insights into
  document embedding generation.
\newblock {\em arXiv preprint arXiv:1607.05368}, 2016.

\bibitem[\protect\citeauthoryear{Le and Mikolov}{2014}]{le2014distributed}
Quoc Le and Tomas Mikolov.
\newblock Distributed representations of sentences and documents.
\newblock In {\em International Conference on Machine Learning}, pages
  1188--1196, 2014.

\bibitem[\protect\citeauthoryear{Lei~Ba \bgroup \em et al.\egroup
  }{2015}]{lei2015predicting}
Jimmy Lei~Ba, Kevin Swersky, Sanja Fidler, et~al.
\newblock Predicting deep zero-shot convolutional neural networks using textual
  descriptions.
\newblock In {\em Proceedings of the IEEE International Conference on Computer
  Vision}, pages 4247--4255, 2015.

\bibitem[\protect\citeauthoryear{Mikolov \bgroup \em et al.\egroup
  }{2013}]{mikolov2013distributed}
Tomas Mikolov, Ilya Sutskever, Kai Chen, Greg~S Corrado, and Jeff Dean.
\newblock Distributed representations of words and phrases and their
  compositionality.
\newblock In {\em Advances in neural information processing systems}, pages
  3111--3119, 2013.

\bibitem[\protect\citeauthoryear{Nikfarjam \bgroup \em et al.\egroup
  }{2015}]{nikfarjam2015pharmacovigilance}
Azadeh Nikfarjam, Abeed Sarker, Karen O’connor, Rachel Ginn, and Graciela
  Gonzalez.
\newblock Pharmacovigilance from social media: mining adverse drug reaction
  mentions using sequence labeling with word embedding cluster features.
\newblock {\em Journal of the American Medical Informatics Association},
  22(3):671--681, 2015.

\bibitem[\protect\citeauthoryear{Noor \bgroup \em et al.\egroup
  }{2008}]{noor2008using}
Abdisalan~M Noor, Victor~A Alegana, Peter~W Gething, Andrew~J Tatem, and
  Robert~W Snow.
\newblock Using remotely sensed night-time light as a proxy for poverty in
  africa.
\newblock {\em Population Health Metrics}, 6(1):5, 2008.

\bibitem[\protect\citeauthoryear{Oshri \bgroup \em et al.\egroup
  }{2018}]{Oshri:2018:IQA:3219819.3219924}
Barak Oshri, Annie Hu, Peter Adelson, Xiao Chen, Pascaline Dupas, Jeremy
  Weinstein, Marshall Burke, David Lobell, and Stefano Ermon.
\newblock Infrastructure quality assessment in africa using satellite imagery
  and deep learning.
\newblock In {\em Proc. of SIGKDD}, pages 616--625. ACM, 2018.

\bibitem[\protect\citeauthoryear{Pang \bgroup \em et al.\egroup
  }{2016}]{pang2016text}
Liang Pang, Yanyan Lan, Jiafeng Guo, Jun Xu, Shengxian Wan, and Xueqi Cheng.
\newblock Text matching as image recognition.
\newblock In {\em AAAI}, 2016.

\bibitem[\protect\citeauthoryear{Perez \bgroup \em et al.\egroup
  }{2017}]{nips_2017_workshop_stefano}
Anthony Perez, Christopher Yeh, George Azzari, Marshall Burke, David Lobell,
  and Stefano Ermon.
\newblock Poverty prediction with public landsat 7 satellite imagery and
  machine learning.
\newblock 11 2017.

\bibitem[\protect\citeauthoryear{Pulse}{2014}]{pulse2014mining}
UN~Global Pulse.
\newblock Mining indonesian tweets to understand food price crises.
\newblock {\em Jakarta: UN Global Pulse}, 2014.

\bibitem[\protect\citeauthoryear{Ramisa \bgroup \em et al.\egroup
  }{2018}]{ramisa2018breakingnews}
Arnau Ramisa, Fei Yan, Francesc Moreno-Noguer, and Krystian Mikolajczyk.
\newblock Breakingnews: Article annotation by image and text processing.
\newblock {\em IEEE TPAMI}, 40(5):1072--1085, 2018.

\bibitem[\protect\citeauthoryear{Sahn and Stifel}{2000}]{sahn2000poverty}
David~E Sahn and David~C Stifel.
\newblock Poverty comparisons over time and across countries in africa.
\newblock {\em World development}, 28(12):2123--2155, 2000.

\bibitem[\protect\citeauthoryear{Sheehan \bgroup \em et al.\egroup
  }{2018}]{sheehan2018learning}
Evan Sheehan, Burak Uzkent, Chenlin Meng, Zhongyi Tang, Marshall Burke, David
  Lobell, and Stefano Ermon.
\newblock Learning to interpret satellite images using wikipedia.
\newblock {\em arXiv preprint arXiv:1809.10236}, 2018.

\bibitem[\protect\citeauthoryear{Signorini \bgroup \em et al.\egroup
  }{2011}]{signorini2011use}
Alessio Signorini, Alberto~Maria Segre, and Philip~M Polgreen.
\newblock The use of twitter to track levels of disease activity and public
  concern in the us during the influenza a h1n1 pandemic.
\newblock {\em PloS one}, 6(5):e19467, 2011.

\bibitem[\protect\citeauthoryear{Smits and
  Steendijk}{2015}]{smits2015international}
Jeroen Smits and Roel Steendijk.
\newblock The international wealth index (iwi).
\newblock {\em Social Indicators Research}, 122(1):65--85, 2015.

\bibitem[\protect\citeauthoryear{Uzkent \bgroup \em et al.\egroup
  }{2019}]{uzkent2019learning}
Burak Uzkent, Evan Sheehan, Chenlin Meng, Zhongyi Tang, Marshall Burke, David
  Lobell, and Stefano Ermon.
\newblock Learning to interpret satellite images in global scale using
  wikipedia.
\newblock {\em arXiv preprint arXiv:1905.02506}, 2019.

\bibitem[\protect\citeauthoryear{Wang \bgroup \em et al.\egroup
  }{2019}]{wang2019learning}
Liwei Wang, Yin Li, Jing Huang, and Svetlana Lazebnik.
\newblock Learning two-branch neural networks for image-text matching tasks.
\newblock {\em IEEE Transactions on Pattern Analysis and Machine Intelligence},
  41(2):394--407, 2019.

\bibitem[\protect\citeauthoryear{Wu \bgroup \em et al.\egroup
  }{2018}]{wu2018image}
Qi~Wu, Chunhua Shen, Peng Wang, Anthony Dick, and Anton van~den Hengel.
\newblock Image captioning and visual question answering based on attributes
  and external knowledge.
\newblock {\em IEEE transactions on pattern analysis and machine intelligence},
  40(6):1367--1381, 2018.

\bibitem[\protect\citeauthoryear{Xie \bgroup \em et al.\egroup
  }{2016}]{Xie2016TransferLF}
Michael Xie, Neal Jean, Marshall Burke, David Lobell, and Stefano Ermon.
\newblock Transfer learning from deep features for remote sensing and poverty
  mapping.
\newblock In {\em Thirtieth AAAI Conference on Artificial Intelligence}, 2016.

\bibitem[\protect\citeauthoryear{Xu \bgroup \em et al.\egroup
  }{2016}]{xu2016cross}
Frank~F Xu, Bill~Y Lin, Qi~Lu, Yifei Huang, and Kenny~Q Zhu.
\newblock Cross-region traffic prediction for china on openstreetmap.
\newblock In {\em Proceedings of the 9th ACM SIGSPATIAL International Workshop
  on Computational Transportation Science}, pages 37--42. ACM, 2016.

\end{thebibliography}
